\documentclass[runningheads]{llncs}
\usepackage[T1]{fontenc}
\usepackage[htt]{hyphenat}
\usepackage{graphicx}
\usepackage{tikz}
\usepackage{wrapfig}
\usepackage{amsmath}
\usepackage{amsfonts}
\usepackage{xcolor}
\usetikzlibrary{positioning}
\usepackage{hyperref}
\usepackage[table]{xcolor}
\usepackage{tabularx}
\usepackage{booktabs}
\usepackage{array}
\usepackage{comment}
\usepackage{bbding}
\usepackage[dvipsnames]{xcolor}
\usepackage[skins,breakable]{tcolorbox}

\renewcommand{\arraystretch}{1.25}

\definecolor{colA}{HTML}{006D77}
\definecolor{colB}{HTML}{83C5bE}
\definecolor{colC}{HTML}{EDF6f9}
\definecolor{colD}{HTML}{FFDDD2}

\definecolor{MyRedBorder}{RGB}{220, 20, 60}
\DeclareRobustCommand{\circled}[1]{\tikz[baseline=(char.base)]{\node[shape=circle,draw=MyRedBorder,text=MyRedBorder,inner sep=0pt,minimum size=4.1mm] (char) {\small #1};}}
\usepackage{enumitem}
\begin{document}

\title{Agentic Orchestration and Orchestration of Agents: Design, Implementation, and Assessment}
\titlerunning{Agentic Orchestration}
%
\author{Stefanie Rinderle-Ma, Juergen Mangler, Johannes Loebbecke, Dominik Voigt, Nataliia Klievtsova, Matthias Ehrendorfer }
\authorrunning{S. Rinderle-Ma et al.}
%
\institute{Technical University of Munich, TUM School of Computation, Information and Technology, Garching, Germany, \email{firstname.lastname@tum.de}}
\maketitle              
\begin{abstract}
Agentic Business Process Management has gained momentum recently. The prospect is that the autonomy of AI agents, i.e., predominantly LLM-based agents, can be balanced with a certain level of robustness, correctness, and traceability through a combination with process technology. In this paper, we provide a classification framework for agentic orchestration options and discuss them along properties such as autonomy, task specificity, reactivity, correctness assurance, traceability, and human involvement. We provide metrics for the quantitative assessment of selected properties and show them through different agentic implementations of a predictive light sensing scenario in CPEE and a blood donation scenario in Langchain. Altogether, this work aims at providing properties and metrics to decide on and assess the design and implementation of agentic orchestrations and orchestration of agents. 

\keywords{Agentic Orchestration  \and Agentic Workflow \and Orchestration of Agents \and Process-Awareness \and Agentic BPM \and LLM Agents  }
\end{abstract}
\section{Introduction}

Recently, Agentic Business Process Management (ABPM) has gained increased attention, resulting in manifestos and perspectives \cite{calvanese_agentic_2026,DBLP:journals/tmis/DumasFLMMRACGFGRVW23,10.1007/978-3-032-02936-2_3}. 
In the ABPM area, agents have been employed for the design and redesign of process and agentic orchestrations \cite{DBLP:conf/bpm/GrohsAER23,DBLP:conf/icsoc/NivonS24,klievtsova_conversational_2026,DBLP:conf/ijcai/KouraniB0A24} as well as for agentic process automation (cf. \cite{DBLP:conf/bpm/MontiLMMR24} and \url{autobpmn.ai}). In doing so, the ABPM approaches adopt a top-down perspective by conceptualizing an explicit model for the agentic orchestration. This, in turn, provides the opportunity to define frames for agentic orchestration ($\mapsto$ framed autonomy), trace their execution, and give assurances regarding the correctness of outputs. Simultaneously, tools such as  \url{crewai.com} or
 \url{langchain.com} have been developed to support the implementation and execution of agent orchestrations. They follow a bottom-up approach where a frame is added to improve results when implementing agent orchestrations.

\begin{figure}[htb!]
    \centering
    \includegraphics[width=0.85\linewidth]{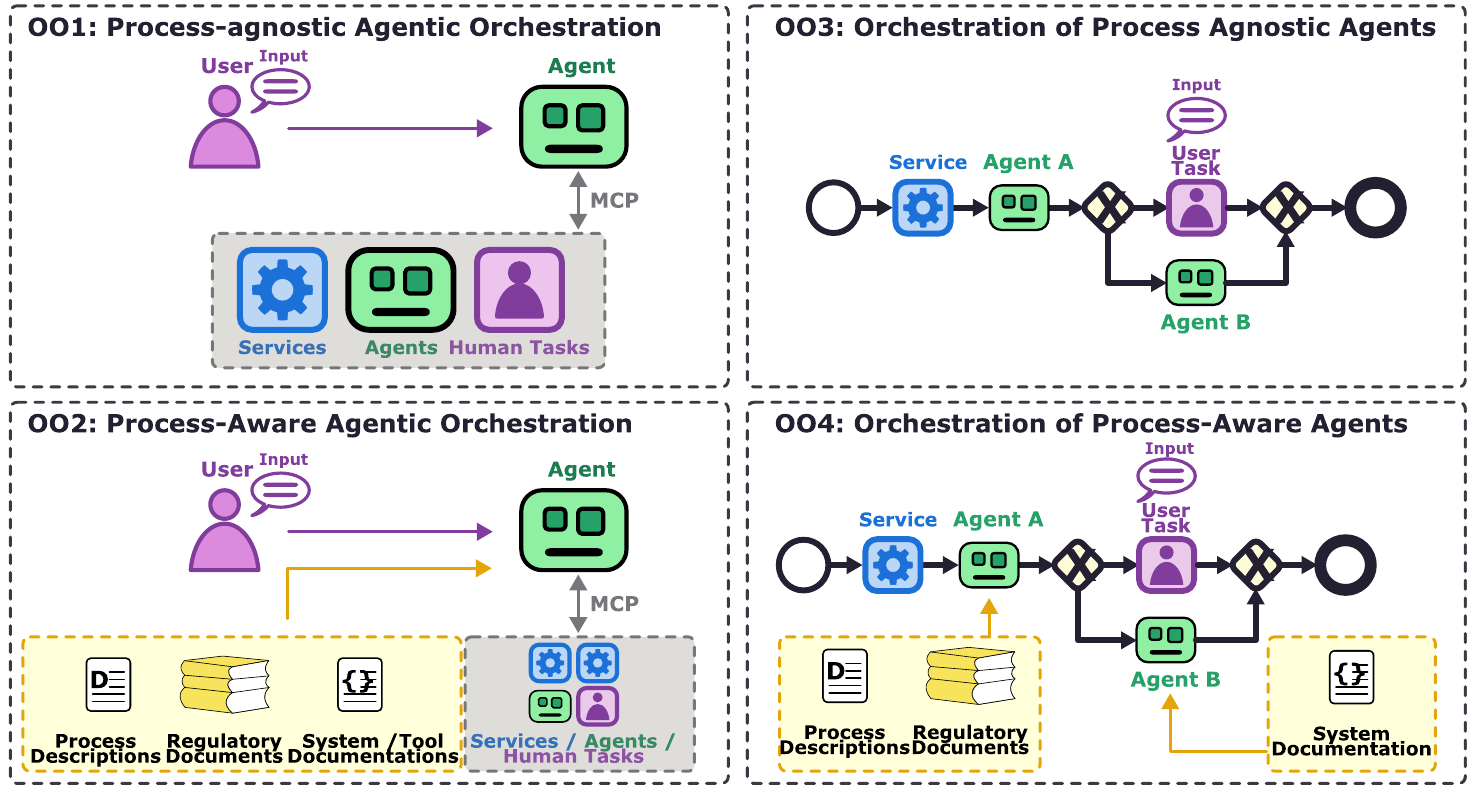}
    \caption{Classification of Options for Agentic Orchestration and Orchestration of Agents }
    \label{fig:matrix}
\end{figure}

However, for both, ABPM and tools, a systematic classification framework to assess different modeling and implementation options for agentic orchestrations is missing, e.g., deciding which functionality is realized by an agent or which agent requires a frame. Choosing an option in an ad-hoc manner can result in insufficient solutions w.r.t. agent properties such as specificity, autonomy, and reactivity on the one side and process properties such as traceability and correctness assurance on the other side. 
Hence, in this work, we combine properties of agents  \cite{DBLP:journals/inffus/SapkotaRK26} with process properties on framed autonomy \cite{calvanese_agentic_2026} as well as properties of human involvement \cite{10.1007/978-3-032-02936-2_3} and assess them along the following \textbf{Orchestration Options (OO)} for realizing scenarios by agents and their orchestrations, illustrated in Fig. \ref{fig:matrix}. Note that the transition between the OOs can be fluid, especially if the process is declarative, resulting in a solution similar to \cite{DBLP:conf/kdd/GuanWC0NSZ24}.

\vspace{-0.18cm}
\begin{description}

   \item[OO1] \textbf{Process-agnostic Agentic Orchestration:} One agent orchestrates multiple work steps, using, for example, protocols such as MCP (Model Context Protocol)\footnote{\url{https://modelcontextprotocol.io/docs/getting-started/intro}} for tool connection and A2A (Agent-to-Agent)\footnote{\url{https://a2aprotocol.ai/}} for agent collaboration. The agent is not aware of any explicit process frame or logic. 
    \item[OO2]  \textbf{Process-Aware Agentic Orchestration:} An agent executes a process orchestration and is (partly) aware of a frame for the execution, stemming from, e.g., regulatory documents or Service Level Agreements (SLAs). No additional frame, ``outside the agent'' is provided. 
    \item[OO3]  \textbf{Orchestration of Process Agnostic Agents:} A (partial) process specification serves as frame in which AI agents are invoked. The agents themselves are process-agnostic. 
       \item[OO4]  \textbf{Orchestration of Process-Aware Agents:} An explicitly defined process orchestrates agents that autonomously execute tasks or sub processes. The agents themselves can also be aware of a frame or process. 
 \end{description}
\vspace{-0.2cm}

The remainder is as follows: Section \ref{sec:classification} defines the orchestration options depicted in Fig. \ref{fig:matrix} and enriches the classification framework with properties from agent  \cite{DBLP:journals/inffus/SapkotaRK26} and process \cite{calvanese_agentic_2026,10.1007/978-3-032-02936-2_3} literature. The properties are qualitatively assessed along a predictive light-sensing scenario. 
Implementations of OO1--OO4 for the predictive light-sensing scenario, using process engine \url{cpee.org}, are provided in Sect. \ref{sec:ligthsensing}. Metrics for quantitatively assessing the properties are added to the framework in Sect. \ref{sec:metrics} and evaluated for the predictive light sensing implementations. Section \ref{sec:blooddonation} implements a complex blood donation scenario using \url{langchain.com} to demonstrate the generalizability of the classification framework. Section \ref{sec:relwork} discusses related work, Sect. \ref{sec:summary} closes with a summary.

\section{Properties of Agentic Orchestrations Options}
\label{sec:classification}

We start by providing a definition for the orchestration options depicted in Fig. \ref{fig:matrix} in order to derive significant properties in the following.

\begin{definition}[Agentic Orchestration]
\label{def:ao}
Let $\cal{A}$ be the set of all LLM agents,  $\cal{H}$ the set of all human agents, $\cal{S}$ the set of all other services, and $\cal{P}$ the set of all process orchestrations described by an orchestration notation such as BPMN or process trees. 
An agentic orchestration $AO:=(A, po, ho, Pa, P)$ comprises: one or several agents $A \subseteq \cal{A}$; a (potentially empty) process orchestration $P$; a certain degree of process-support $po: P \mapsto \mathbb{N}_0$, where $po = 0$ implies that the level of process-support is unknown (i.e., hidden in the model logic), while $po>0$ implies that more orchestration logic is handled explicitly~\footnote{We propose two metrics to calculate the degree of explicit process support in Sec.~\ref{sec:metrics}}; and a certain level of human involvement $ho \in$ \{implicit, explicit\} where implicit refers to monitoring and input provision and explicit refers to invocation of human agents. Moreover, let $pa: A \mapsto \{aware, agnostic\}$ and $Pa=\{(a,pa(a))|a \in A\}$ determine for each agent in AO whether it is process-aware or agnostic. 
If po=explicit, then $P \in \cal{P}$ describes the process orchestration based on $A \cup H \cup S$; if ho=explicit, $H \subseteq \cal{H}$ denotes the set of involved human agents (otherwise: $H=\emptyset$ ); $S \subseteq \cal{S}$ if other services are involved (otherwise: $S=\emptyset$). 
 \end{definition}

Consider the orchestration options depicted in Fig. \ref{fig:matrix}: 
Given two Agents $A$ and $B$, an AO under orchestration option OO could be defined as AO$_{OO1}$ = (\{$A$\}, 0, implicit, \{($A$, agnostic)\}, $\emptyset$), AO$_{OO2}$ = (\{$A$\}, 0, implicit, \{($A$, aware)\}, $\emptyset$), AO$_{OO3}$ = (\{$A$, $B$\}, $po>0$, explicit, \{($A$, agnostic), ($B$, agnostic)\}, P on \{$A$, $B$, Service, User Task\}), and  AO$_{OO4}$ = (\{$A$, $B$\}, $po>0$, explicit,  \{($A$, aware), ($B$, aware)\}, P on \{$A$, $B$, Service, User Task\}).
The properties of an AO under orchestration option OO can be derived from properties of the involved agents, processes, and human involvement.

\noindent\textbf {Agent Properties}
comprise autonomy, task specificity, and reactivity  \cite{DBLP:journals/inffus/SapkotaRK26}. 
\textsl{Autonomy} 
refers to the ability to take decisions with minimal human intervention. The autonomy of an agent can be restricted by a frame (i.e., framed autonomy \cite{calvanese_agentic_2026}). The frame can be realized using rules in a declarative process description or by imperative process fragments. The autonomy of an agent could be measured, for example, by the number of decisions an agent can take. However, as agent behavior is often black box, it is hard to estimate the full decision space. Autonomy is connected to \textsl{task specificity} of agents which refers to the granularity an agent is operating on. For OO3 and OO4, task specificity is mapped to the granularity of the process tasks which invoke the agents, i.e., we can describe the degree of process-support by analyzing how much orchestration logic is handled explicitly. As such, in Sect. \ref{sec:metrics}, we propose and investigate Cyclomatic complexity \cite{mccabe1976complexity} and ABC Complexity \cite{fitzpatrick1997applying} as metrics to assess task specificity. \textsl{Reactivity} refers to the ability of an agent to respond to environment changes and can be most probably only observed during runtime. We propose to use reactivity-speed as a metric to partially capture reactivity in Sect. \ref{sec:metrics}.

\noindent\textbf{Process Properties} refer to properties that are induced by frames and processes according to \cite{calvanese_agentic_2026} and include (among others) correctness assurance and traceability.  
\textsl{Correctness assurance} refers to the ability to give guarantees for an achieved state or outcome. It can be assessed using precision, recall, and F1 score. \textsl{Traceability} refers to the ability to understand \underline{why} a decision in the agentic orchestration was taken and how a certain state or outcome was achieved. 
We propose to assess traceability based on the event logs generated by executing the agentic orchestration compared to all events that have actually occurred.

\noindent\textbf {Human Involvement Properties} include, according to \cite{10.1007/978-3-032-02936-2_3}, human-agent collaboration and balance as well as involvement of humans.  
\textsl{Human-agent collaboration and balance} means to establish clear responsibilities and leverage on the strengths of both, agent and human. Moreover, this property refers to enabling humans to act in case of agent failure. \textsl{Human-agent involvement} can range from human as peer of the agent to human as its supervisor. For OO1 and OO2, the human involvement refers to the providing input to the agent and to human tasks potentially being invoked by the agent. For OO3 and OO4, the explicit process definition contributes to a clear picture on which tasks are invoking an agent, service, or human task. A simple metric could be to calculate some ratio between agent and human tasks. Moreover, data flow dependencies between agent and human tasks can be analyzed. Due to space, we will not provide quantitative metrics for human involvement in Sect. \ref{sec:metrics}, but in future work.

In the following, we illustrate each OO based on a light sensing agent orchestration similar to the scenarios described in \cite{DBLP:journals/patterns/AnZZY26,DBLP:journals/inffus/SapkotaRK26}:

\begin{tcolorbox}[title=BASIC (OO1),
title filled=false,
colback=NavyBlue!5!white,
colframe=NavyBlue!75!black,
breakable, left=2pt, right=2pt, top=2pt,bottom=2pt]
Ambient light conditions are constantly measured by light sensors. Depending on the light intensity, the light in the room is switched on, off, or maintained. 
\end{tcolorbox}

BASIC can be realized by OO1, i.e., by process-agnostic agentic orchestration where the agent orchestrates and executes services and other agents, possibly via protocols such as MCP, and can take input from users. No process logic or frame is defined, i.e., AO$_{OO1}$ = (\{Agent\}, 0, implicit, \{(Agent, agnostic)\}, $\emptyset$)

\noindent\textsl{\textcolor{NavyBlue}{BASIC~(OO1)~\HandRight~Agent Properties:}} 
As the agent in OO1 acts independently of any frame or process, its autonomy and reactivity are high, and the task specificity might be medium to low. Due to the low task specificity, the functionality could be also realized by a service instead of an agent, 
especially if the rules for switching the light on/off are fixed.  

\noindent\textsl{\textcolor{NavyBlue}{BASIC~(OO1)~\HandRight~Process Properties:}} 
As the agent behavior is black box, OO1 is low in traceability, tractability, and correctness assurance, i.e., we can neither provide information on how an outcome was achieved nor give any assurances on the outcome correctness or quality. In BASIC, this means that we can observe the outcome, i.e., light on or off, but not reproduce why. 

\noindent\textsl{\textcolor{NavyBlue}{BASIC~(OO1)~\HandRight~Human Properties:}} Human involvement is confined to possible prompting.   
BASIC does neither require human oversight of internal steps nor involves human work/actions. 

Assume now that a condition on switching the light off after a certain period of time is added to BASIC, resulting in FRAME:

\begin{tcolorbox}[title=FRAME (OO2),
title filled=false,
colback=NavyBlue!5!white,
colframe=NavyBlue!75!black,
breakable,left=2pt, right=2pt, top=2pt,bottom=2pt]
Ambient light conditions are constantly measured by light sensors. Depending on the light intensity, the light in the room  switched on, switched off, or maintained.  If the light has been on for 10 hours, it should be switched off.
\end{tcolorbox}

FRAME could be realized by OO2 where the agent is aware of a frame given by, e.g., process descriptions, regulatory documents, or documentations. The frame can be directly included in the prompt or extracted from additional sources by the agent, i.e., AO$_{OO2}$ = (\{Agent\}, 0, implicit, \{(Agent, aware)\}, $\emptyset$).  

\noindent\textsl{\textcolor{NavyBlue}{FRAME~(OO2)~\HandRight~Agent Properties:}} The autonomy and reactivity might be restricted when compared to OO1 due to the imposed frame. Task specificity remains similar. 
 
\noindent\textsl{\textcolor{NavyBlue}{FRAME~(OO2)~\HandRight~Process Properties:}} 
As OO2 is aware of a frame, this results in higher traceability and tractability while the autonomy of the agent is still high. The frame might result in reduced reactivity, especially if it restricts the action space to react on changes in the environment. For FRAME, this might be the case if the outside light conditions become suddenly very dark, e.g., due to a thunderstorm, but the lighting has been on for $10$ hours. Correctness assurance might be the same or a bit higher than for OO1, depending on the influence of the frame on the outcome of the agentic orchestration.

\noindent\textsl{\textcolor{NavyBlue}{FRAME~(OO2)~\HandRight~Human Properties:}}  Human involvement is confined to possible prompting.   
OO2 does neither include human oversight of internal steps and/or human interaction nor involves human work/actions.

Assume that FRAME is augmented with an initial collection of user preferences, resulting in FRAME-USER:
\begin{tcolorbox}[title=FRAME-USER (OO3),
title filled=false,
colback=NavyBlue!5!white,
colframe=NavyBlue!75!black,
breakable,left=2pt, right=2pt, top=2pt,bottom=2pt]
User preferences regarding the lighting in the building are collected. 
Ambient light conditions are constantly measured by light sensors. Depending on the light intensity and user preferences, the light in the room is switched on, switched off, or maintained.  If the light has been on for 10 hours, it should be switched off.
\end{tcolorbox}

We can realize FRAME-USER using OO3 with an agent that handles the user preferences and is process-agnostic. The frame, i.e., switching off the light after $10$ hours, is realized by embedding the agent into a process orchestration.

\noindent\textsl{\textcolor{NavyBlue}{FRAME-USER~(OO3)~\HandRight~Agent and Process Properties:}}
For OO3, the frame realized by the orchestration can provide increased correctness assurance. In the example, it can be ensured that the light is switched off for sure after 10 hours. Deciding which parts of the frame are realized ``outside'' the agent is related to deciding on the decision space of the agent and hence on its autonomy, reactivity, and task specificity. As the agent itself is process-agnostic and rather black box, the behavior realized by the agent lacks traceability and tractability. 

\noindent\textsl{\textcolor{NavyBlue}{FRAME-USER~(OO3)~\HandRight~Human Properties:}} 
OO3 could be desirable for scenarios that  do not require human oversight of certain steps, but involve human work/actions.

We now extend the light sensing example to a smart building example by additionally collecting user preferences regarding the temperature and turning the heating on/off accordingly:

\begin{tcolorbox}[title=ORCHESTRATE (OO4),
title filled=false,
colback=NavyBlue!5!white,
colframe=NavyBlue!75!black,
breakable,left=2pt, right=2pt, top=2pt,bottom=2pt]
<FRAME-USER>. In parallel, the room temperature is measured and the heater is turned on/off based on collected user preferences.
\end{tcolorbox}

ORCHESTRATE can be realized by OO4 where (multiple) agents are invoked by a process orchestration (frame) and are process-aware themselves.

\noindent\textsl{\textcolor{NavyBlue}{ORCHESTRATE~(OO4)~\HandRight~Agent and Process Properties:}}
OO4 refers to the (framed) orchestration of process-aware AI agents, resulting in actions spaces of varying autonomy of multiple agents, higher correctness assurance, and higher traceability and tractability.

\noindent\textsl{\textcolor{NavyBlue}{ORCHESTRATE~(OO4)~\HandRight~Human Properties:}} 
OO4 could be desirable for scenarios that  include human oversight of internal steps and/or human interaction as well as involve human work/actions.

Note that the options for realizing the different scenario variants BASIC, FRAME, FRAME-USER, and ORCHESTRATE are fluid, i.e., different options can be chosen for the different scenarios. It would be also possible to realize the scenario entirely without an  agent, i.e., by a process orchestration.

\section{Design and Implementation of Agentic Light Sensing}
\label{sec:ligthsensing}

This section designs, implements, and simulates OO1--OO4 (cf. Fig. \ref{fig:matrix}) for the predictive light sensing scenario introduced in Sect. \ref{sec:classification}: we consider a cyber-physical scenario which measures lighting based on lux ($\frac{Lumens}{Area}$), movement, and occupancy count inside a room and adjusts the lumen according to the sensor values, user inputs and/or specifications (which are either integrated into the frame, process orchestration, or can be considered via using MCP tools) using an LLM agent. We create an artificial dataset consisting of x measurement events, for which occupancy count and movement is assumed minimal during night hours. 
For each measurement we calculate a range of ground truth lumen values according to the set of rules 
in Tab.~\ref{tab:lumen_rules} using the assumption of a single lamp that can produce any lumen value between 0 and 5000 and is positioned at 3 meters above the floor with a circular beam angle of $60^{\circ}$ that uniformly covers the entire room\footnote{Lux formula: 
$\text{Lux}
=
\frac{\text{Lumens}}{2\pi(3)^2\left(1-\cos\left(30^\circ\right)\right)}
=
\frac{\text{Lumens}}{18\pi\left(1-\frac{\sqrt{3}}{2}\right)}
\approx
\frac{\text{Lumens}}{7.5761}$
}.
To simulate real sensors, the measurement events and a lamp endpoint are provided via REST interfaces, which are connected to an LLM agent via MCP. We experimented with the following LLMs, i.e., \texttt{mistralai/Ministral-3-14B-Reasoning-2512}, \texttt{google/gemma-4-31B-it}, \texttt{Qwen/Qwen3.6-35B-A3B}, and \texttt{qwen-35-35b-coding}. Details on the implementation can be found on GitHub\footnote{\url{https://anonymous.4open.science/r/AgenticProcessOrchestration/README.md}}.

\begin{table}[htb!]
\centering
\scriptsize
\renewcommand{\arraystretch}{1.2}
\setlength{\tabcolsep}{6pt}
\begin{tabularx}{\linewidth}{
  >{\centering\arraybackslash}p{0.055\linewidth}
  >{\raggedright\arraybackslash}p{0.50\linewidth}
  >{\raggedright\arraybackslash}X
}
\hline
\rowcolor{gray!15}
\textbf{Prio} & \textbf{Condition} & \textbf{Consequence} \\
\hline
\rowcolor{blue!12}
0 & User input present & React to user input request \\
1 & Occupancy $< 1$ and current lumen == 0 & Do not change lamp lumen \\
2 & Occupancy $< 1$ and current lumen =! 0 & Set lumen to $0$ \\
3 & Night ($23$--$6$) and movement detected & Target $50$ lux \\
4 & Night ($23$--$6$) and no movement & Set lumen to $0$ \\
5 & Morning ($6$--$9$) and occupancy $\ge 1$ & Target $100$--$200$ lux \\
6 & Midday ($9$--$14$) and occupancy $\ge 1$ & Target $300$--$500$ lux \\
7 & Afternoon ($14$--$18$) and occupancy $\ge 1$ & Target $200$--$300$ lux \\
8 & Evening ($18$--$23$) and occupancy $\ge 1$ & Target $200$--$300$ lux \\
9 & Else: (default deny) & Do not change lamp lumen \\
\hline
\end{tabularx}
\caption{Rules considered as the ground truth for the lighting scenario using default deny semantics. User-input rule 0 is only applicable in the LLM-based variants.}
\label{tab:lumen_rules}
\end{table}
\vspace{-0.6cm}

For implementing OO1, the agent is not framed and contains minimal context information through the following prompt which refines BASIC in Sect. \ref{sec:classification}: 

\begin{tcolorbox}[title=PROMPT OO1 ($\rightarrow$ BASIC),
title filled=false,
colback=NavyBlue!5!white,
colframe=NavyBlue!75!black,
breakable, left=2pt, right=2pt, top=2pt,bottom=2pt]
 Adjust the lighting based on standard guidelines for brightness during day/night and sleep/being awake. Also consider that the light should be turned on whenever it is dark and I want to go somewhere! Make sure this is done for the next 72 seconds which corresponds to 12 hours in simulated time (check for the conditions every second - i.e., every 10 minutes in simulated time)! Log the tools you use along with their parameters (apart from the logging tool itself)!
 \end{tcolorbox}

For OO2, the agent is not framed by a process, but aware of a frame provided to the agent via context information through the following prompt: 

\begin{tcolorbox}[title=PROMPT OO2 ($\rightarrow$ FRAME),
title filled=false,
colback=NavyBlue!5!white,
colframe=NavyBlue!75!black,
breakable, left=2pt, right=2pt, top=2pt,bottom=2pt]
Adjust the lighting based on the following rules: Do not change the lumen of the lamp unless any of the following rules hold: (1) If there is user input react to user input, (2) If the occupancy is less than 1 then lower lumen to 0, (3) In the night (23-6) if there is movement aim for 50 lux, (4) In the night (23-6) if there is no movement lower lumen to 0, (5) In the morning (6-9) if occupancy is greater or equal 1 then aim for 100-200 lux, (6) during midday (9-14) if occupancy is greater or equal 1 then aim for 300-500 lux, (7) In the afternoon (14-18) if occupancy is greater or equal 1 aim for 200-300 lux, and (8) In the evening (18-23) if occupancy is greater or equal 1 then aim for 200-300 lux! Make sure this is done for the next 72 seconds which corresponds to 12 hours in simulated time (check for the conditions every second - i.e., every 10 minutes in simulated time)! Log the tools you use along with their parameters (apart from the logging tool itself)!
\end{tcolorbox}

\begin{figure}[tb!]
    \centering
    \includegraphics[width=0.8\linewidth]{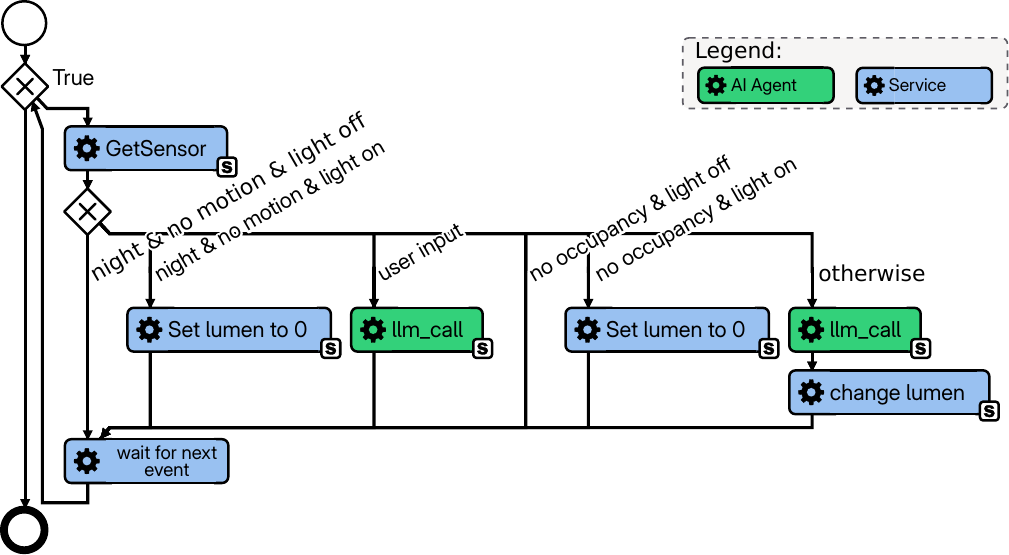}
    \caption{Orchestration of agents for lighting scenario implementation (OO3 and OO4)}
    \label{fig:light_implem_oo3}
\end{figure}

For the implementation of OO3 and OO4, the agent is framed by a process illustrated in Fig.~\ref{fig:light_implem_oo3} that ensures some rules are deterministically enforced, either provided a simple prompt for OO3 or a complex prompt for OO4 (without the deterministically enforced rule). Note that here the scenario partly deviates from the simplistic descriptions in Sect.~\ref{sec:classification}.

\begin{tcolorbox}[title=PROMPT OO3,
title filled=false,
colback=NavyBlue!5!white,
colframe=NavyBlue!75!black,
breakable, left=2pt, right=2pt, top=2pt,bottom=2pt]
It is currently \{time\}, the lamp is currently at \{current\} and the current ambient lux is at \{ambient\}. Return a lumen value based on standard guidelines for brightness during day/night and sleep/being awake. Also consider that the light should be turned on whenever it is dark and I want to go somewhere! Return just the lumen value as an int.
\end{tcolorbox}

\begin{tcolorbox}[title=PROMPT OO4,
title filled=false,
colback=NavyBlue!5!white,
colframe=NavyBlue!75!black,
breakable, left=2pt, right=2pt, top=2pt,bottom=2pt]
It is currently \{time\}, the lamp is currently at \{current\} and the current ambient lux is at \{ambient\}. Return the lumen value that a lamp should take according to the following rules: (1) If it is night, turn of the Light if it is running. (2) If it is morning aim for 100-200 lux, (3) If it is midday aim for 300-500 lux, (4) If it is afternoon aim for 200-300 lux, and (5) If it is evening aim for 200-300 lux! Return just the Lumen Value as an Int.
\end{tcolorbox}

Finally, to provide a better comparison we create a purely deterministic (manually modeled) version of OO4 denoted as $\mathbb{GT}$, which correctly enforces rules 1$\to$9 (cf. \footnote{\scriptsize\url{https://cpee.org/hub/server/Papers.dir/AgenticFundamentals.dir/005.xml/open?stage=view}}).

\noindent\textbf{Evaluation and Conclusion:} The design and implementation of the scenarios show the feasibility of the framework presented in Fig. \ref{fig:matrix}. Moreover, the following qualitative observations have been made: (1) higher autonomy is often assumed for agent solutions, but the actual information and action space of agents is clearly restricted through the connected tools, which have to be connected like in a traditional workflow orchestration. (2) Any temporal constraints are challenging to enforce, as the reasoning times of LLMs are unpredictable, so they require frames/tools to enforce temporal relationships. (3) Relying on agents for logging is generally not recommended, even when logging tools are provided. (4) It is challenging to control the LLM context window for long workflows. (5) The less framed the orchestration the more challenging it is to debug, due to the different software layers which require extensive boilerplate code to ensure messages are correctly propagated (i.e., LLM-Agent calls MCP tool, which calls a REST service, which connects to a MQTT server, which connects to a sensor). (6) The less framed the orchestration, the more confounding variables via textual descriptions are added. (7) An unframed orchestration does not require translating textual requirements into a formal language.

\section{Metrics for Orchestration Option Assessment}
\label{sec:metrics}

\begin{figure}[htb!]
    \centering
    \includegraphics[width=\linewidth]{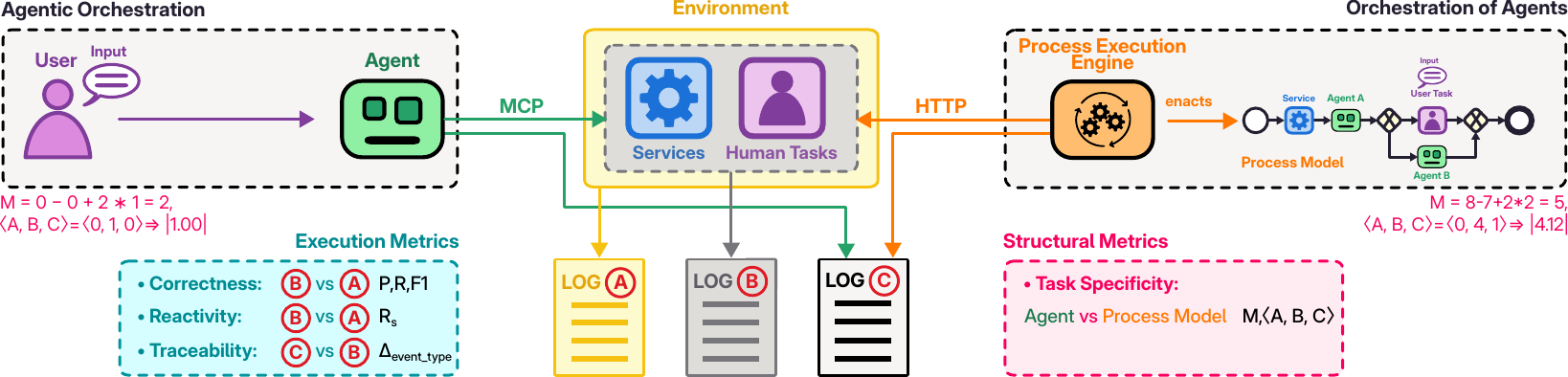}
    \caption{Overview of Proposed Metrics}
    \label{fig:metrics_overview}
\end{figure}

Section \ref{sec:ligthsensing} discusses the agent, process, and human properties introduced in Sect.~\ref{sec:classification} for the lighting scenario in a qualitative way. 
To additionally enable their quantitative assessment along the metrics in Fig.~\ref{fig:metrics_overview}, we propose first metrics to capture task specificity, correctness, reactivity, and traceability, whereas autonomy and tractability are subject to future work. Figure~\ref{fig:metrics_overview} shows how the scenarios (i.e., Agentic Orchestration as in OO1 and OO2 as well as Orchestration of Agents as in OO3 and OO4) are implemented and on which data the calculation of metrics is based. \circled{A} represents the ground truth (manually created based on rules 1$\to$9), \circled{B} the logs created by the invoked (either by the agent in the case of Agentic Orchestration or by the Process Execution Engine in the case of Orchestration of Agents) services (e.g. check the light sensor or change the lumen), and \circled{C} the logs created by the orchestrator (i.e., either the Agent or Process Execution Engine). Execution Metrics are calculated based on a comparison between \circled{A}, \circled{B}, and \circled{C} while Structural Metrics are based on analyzing the specification of the underlying scenario. We present proposed metrics with associated calculation methods and their values based on $10$ runs of the implemented Agentic Light Sensing Scenario (Sect.~\ref{sec:ligthsensing}) using the \texttt{google/gemma-4-31B-it} model below.

\noindent \textbf{Task Specificity:} As illustrated in Fig.~\ref{fig:metrics_overview}, we propose two structural measures for task specificity of agent orchestrations based on well-established software complexity measures, i.e., the Cyclomatic complexity $M$~\cite{mccabe1976complexity} and the ABC Complexity $\langle ABC\rangle$~\cite{fitzpatrick1997applying}. The intuition is to consider a (potentially imagined) process that surrounds the agent orchestration: In the case of orchestrations in OO1/OO2, the agent(s) are not included in a surrounding process, and as such the task specificity is minimal, while if the agents are included in a workflow (such as in a process orchestration of style OO3/OO4), the task specificity increases depending on how much agent logic is moved to the process logic. As the amount of logic in the process can be captured using software complexity metrics, they can be used to compare the task specificity of different agent orchestrations for a process/workflow. An overview of different example orchestrations with the respective measures is in Tab.~\ref{tab:oe-metrics} and calculation scripts are provided on GitHub$^5$.

The Cyclomatic complexity of a process is defined as $M := E-N+2P$ where $E$ refers to the number of edges and $N$ to the number of nodes on the longest path resp., and $P$ refers to the number of connected components. For $M$ it is irrelevant whether the start/end nodes as well as its connectors are considered for the process model as they add a single node/edge each and thus cancel out. The same applies regarding whether every decision/parallel leads back to a decision/parallel (i.e., block structuredness) or directly to the next node. For OO1 and OO2, i.e., a single agent without any process frame, the agent is the only connected component; thus: $M = 0-0+2*1 = 2$. For OO3 and OO4, the task specificity increases depending on how much logic is explicitly included in the process model vs how much logic is implicitly left to the agent. E.g., in Fig.~\ref{fig:metrics_overview}, the orchestration model has 8 Edges, 7 Nodes on the longest path and 2 agents (software components), i.e., $M = 8-7+2*2 = 5$.

ABC complexity has been proposed as a metric that can be calculated in a single pass over the data structure ($O(N)$) while presenting better insights and being more adaptable than other simple metrics such as lines of code. ABC metric is defined as the vector $\langle A,B,C\rangle$ where $|A|$ is the amount of Assignments, $|B|$ the amount of Branches, and $|C|$ is the amount of Conditionals, with the associated ABC score defined as its scalar $|ABC| = \sqrt{A^2+B^2+C^2}$. As branches in this context are explicit forward program branches out of scope of the original program, we can interpret $B$ as the set of activities and subprocesses (including agents), while $C$ refers to the set of exclusive, parallel, and loop operators. Regarding assignments $A$, we consider only script tasks, which contain (usually short) data object computations, as part of $A$. This also enables using different weights depending on the scenario. For example, it can make sense to consider activities/subprocesses as making the orchestration less task specific than moving logic into the process model by including it via operators. In this case a weighted $|ABC_w|$ score such as $|ABC_w| = \sqrt{A^2+0.8*B^2+C^2}$ could be appropriate. Similarly the metric can also be extended to enable different weights depending on the type of activity. For example a $ABCD$ metric, where $D$ contains all agent tasks can help highlight the degree of autonomy in addition to comparing task specificity.

Both measures can be applied analogously to compare the task specificity as defined in current frameworks like crewAI flows or AgentSDK workflow-patterns. However, it is important to note that $M$ and $ABC$ provide only a comparative measure between flows/orchestrations of the \emph{same} process/workflow, and do not have any semantic meaning when considering a single orchestration in isolation or when comparing two orchestrations of different processes/workflows.

\noindent \textbf{Correctness:} For correctness, we consider standard execution metrics as illustrated using \circled{A} and \circled{B} in Fig.~\ref{fig:metrics_overview}: We consider a true positive (TP) as an event $e$ that was executed and required according to some set of rules $\mathbf{R}$ (i.e., $e\in$ \circled{B} $\wedge \ e \in$ \circled{A}) and a false positive (FP) as an event that was executed, but not required according to $\mathbf{R}$ ($e \in$ \circled{B} $\wedge \ e \notin$~\circled{A}). Regarding false negative (FN) we differentiate $FN$ and \emph{time-sensitive} $FN_t$: $FN$ only considers events that were not executed, but required according to $ \mathbf{R}$ ($e \notin$ \circled{B} $\wedge \ e \in$ \circled{A}) and a previously observed event $e'$ ($ e \notin$ \circled{A} $ \wedge \ \exists e' \in$ \circled{B}$ \exists r \in \mathbf{R}: r = e' \implies e)$, while $FN_t$ follows the standard $e$ exists in \circled{A} but not in \circled{B}. This differentiation is added, as $FN_t$ considers events that should have been observed by the orchestration, but were missed as the agent was still processing a previously observed event as false negatives, i.e., $FN_t$ depends on the available computing power. Meanwhile, $FN$ can be seen as \emph{ideal} FN value, which assumes infinite computing power, such that the agents do not miss any events. As agents are expensive/slow in practice we capture the high amount of missed events in time-critical applications using $FN_t$. As such, we evaluate correctness using precision $P$, recall $R$, time-sensitive recall $R_t$, and the combined $F1$ and $F1_t$ scores calculated as: $P=\frac{TP}{TP+FP}$, $R=\frac{TP}{TP+FN}$, $R_t=\frac{TP}{TP+FN_t}$, $F1=2*\frac{P*R}{P+R}$, $F1_t=2*\frac{P*R_t}{P+R_t}$. 

\noindent \textbf{Reactivity:} We capture the reactivity-speed $R_s$, defined as the average time between an event that requires a consequence event, according to $\mathbf{R}$. 

\noindent \textbf{Traceability:} We consider traceability as the correctness of the log created by orchestrators \circled{C} compared with the actual execution log~\circled{B}. While this could theoretically be captured using $P$, $R$, and $F1$ scores, in practice these are challenging to calculate, as event matching between the two logs, which use different timestamps/IDs for the same event, is required. We instead compare the amounts of events per type, captured using the mean and standard deviation of the difference between \circled{B} and \circled{C} in Tab.~\ref{tab:oe-metrics} for every event type: ``check environment'' ($\Delta_{env}$) and ``change lumens'' ($\Delta_{lum}$).

\begin{table}[htb!]

\centering
\scriptsize
\setlength{\tabcolsep}{4pt}
\renewcommand{\arraystretch}{1.15}
\begin{tabular}{lccccc}
\toprule
\textbf{Metric} & \textbf{OO1} & \textbf{OO2} & \textbf{OO3} & \textbf{OO4} & \textbf{$\mathbb{GT}$} \\
\midrule
$P$    & $0.52\pm0.20$ & $0.61\pm0.13$ & $0.69\pm0.14$ & $0.71\pm0.14$ & $1.0\pm0.0$ \\
$R$    & $0.39\pm0.34$  & $0.58\pm0.27$ & $1.0\pm0.0$ & $1.0\pm0.0$ & $1.0\pm0.0$ \\
$R_t$  & $0.16\pm0.19$ & $0.25\pm0.19$ & $0.18\pm0.05$ & $0.29\pm0.10$ & $1.0\pm0.0$ \\
$F1$   & $0.40\pm0.28$  & $0.55\pm0.17$ & $0.81\pm0.10$ & $0.82\pm0.11$ & $1.0\pm0.0$ \\
$F1_t$ & $0.21\pm0.21$ & $0.33\pm0.18$ & $0.28\pm0.07$ & $0.41\pm0.11$ & $1.0\pm0.0$ \\
$R_s$  & $5.46\pm3.16$ & $4.30\pm1.45$ & $5.37\pm2.99$ & $4.41\pm2.04$ & $0.12\pm0.01$ \\
$\Delta_{lum}$ & $6.1\pm13.10$ & $1.3\pm4.24$ & $0.0\pm0.0$ & $0.0\pm0.0$ & $0.0\pm0.0$ \\
$\Delta_{env}$ & $35.4\pm31.92$ & $106.1\pm118.21$ & $0.0\pm0.0$ & $0.0\pm0.0$ & $0.0\pm0.0$ \\
\bottomrule
\end{tabular}
\caption{Correctness and Reactivity Metric comparison of OO1--$\mathbb{GT}$ (cf. Sect. \ref{sec:ligthsensing}).}
\label{tab:oe-metrics}
\end{table}
\vspace{-0.7cm}

\noindent\textbf{Results:} Analyzing Tab.~\ref{tab:oe-metrics}~and~\ref{tab:structural-metrics}, we identify that $M$ and $ABC$ correctly capture task specificity as expected, i.e., an orchestration where the agent is only responsible for more specific cases has higher $M$ and $ABC$ scores. Comparing them we find the $ABC$ metric more useful and adaptable while $M$ can capture the basic relationship in a single value. Regarding correctness, we see how context information always improves $P$ and $R$, while including a frame also requires additional context information to improve correctness. Regarding reactivity, we cannot identify any relevant information
because $R_s$ depends on too many other factors such as model load, which are not accounted for.
Analyzing traceability we identify that agents do not generate reliable event logs when provided logging tools ($\Delta>0$ for OO1 and OO2). As such, we argue that logging functionalities should always be handled through middleware instead, regardless of chosen OO. 

\begin{table}[tbh!]
\centering
\scriptsize
\setlength{\tabcolsep}{6pt}
\renewcommand{\arraystretch}{1.15}
\begin{tabular}{lccccc|cc}
\toprule
\textbf{Metric} & \textbf{OO1} & \textbf{OO2} & \textbf{OO3} & \textbf{OO4} & \textbf{$\mathbb{GT}$} & \textbf{CN} & \textbf{COO4} \\
\midrule
$M$ & 2 & 2 & 13 & 13 & 17 & 2 & 12 \\
$\langle A,B,C\rangle$ & $\langle0,1,0\rangle$ & $\langle0,1,0\rangle$ & $\langle0,4,2\rangle$ & $\langle1,5,2\rangle$ & $\langle8,4,5\rangle$ & $\langle0,1,0\rangle$ & $\langle6,7,4\rangle$ \\
$|ABC|$ & 1.00 & 1.00 & 4.47 & 5.48 & 10.25 & 1.00 & 10.05 \\
\bottomrule
\end{tabular}
\caption{Structural Metric comparison of OO1--$\mathbb{GT}$ (Sect. \ref{sec:ligthsensing}). and CN-COO4 (Sect.~\ref{sec:blooddonation})}
\label{tab:structural-metrics}
\end{table}
\vspace{-0.7cm}

\noindent\textbf{Threats to validity:} We only present the results of initial experiments to illustrate our proposed metrics which are subject to  several glaring threats of validity such as the lack of a comparison of different prompts, potential overfitting in prompt engineering, confounding factors such as model load and network delays, and the chosen models. It is especially important to highlight that for the ground truth values we did not consider user inputs, which is the primary reason for including an agent in this scenario.

\section{Complex Real World Scenario: Blood Donation}
\label{sec:blooddonation}

\begin{figure}
    \centering
    \includegraphics[width=0.85\linewidth]{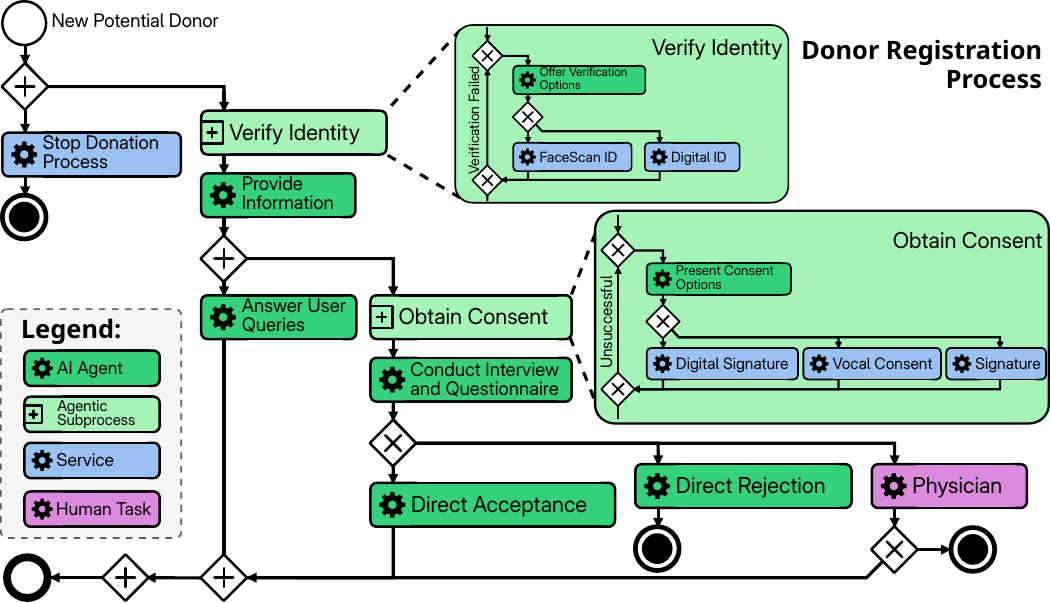}
    \caption{Process-framed Agentic Blood Donor Registration based on the Blood Directive (2002/98/EC) \& Blood Guide$^6$, May 21. 2026).}
    \label{fig:donor_registration}
\end{figure}

We finally illustrate a complex real world scenario, based on guidelines for the use and quality assurance of blood components given in the European Blood Guide\footnote{\url{https://www.edqm.eu/en/blood-guide}}.
In this guide, the \emph{donor registration} process is strictly regulated and consists both of activities that can be automated, such as the identify verification and the provision of relevant information, as well as human tasks where the physician in charge of the blood donation has to make the final eligibility decision in specific cases identified through a questionnaire. An OO4 framed/aware agent orchestration of this process is illustrated in Fig.~\ref{fig:donor_registration}.

A naive orchestration of this scenario would be a single agent with connected tools and the task to \texttt{complete the registration of a blood donor}. In this scenario, such an unframed/unaware orchestration is unacceptable as there are strict execution and documentation requirements enforced by the blood directive. Providing context information, by e.g., uploading the Blood Guide to a vector database for a Retrieval Augmented Generation Agent orchestration would likely improve the behavior, but still entails unacceptable risks. Specifically essential tasks such as a donor always having the opportunity to retract their consent and abort the donation should be deterministically enforced. As such a framed solution (i.e., Fig.~\ref{fig:donor_registration}), where the stopping is enabled through a service in parallel with different agents orchestrated in a workflow is appropriate in this case. A further question then becomes which task specificity is appropriate for this solution. For example, the identity verification and the obtaining of consent, can be both pure single agents with connected tools, purely deterministic, agents with additional context information or agentic sub-processes, where a frame surrounds the agent to restrict which choices can actually be taken. However, there are also tasks where a purely deterministic solution is unsuited, as the constraints enforce that potential donors have the ability to ask questions regarding the nature of blood donations, which is suitable for a LLM agent. We implement the naive (CN in Tab.~\ref{tab:oe-metrics}) and a OO4 framed-orchestration solution (COO4 in Tab.~\ref{tab:oe-metrics}) using langchain\footnote{\url{https://github.com/langchain-ai/langchain}} and various simple simulators for different service endpoints, to illustrate how the cyclomatic complexity and ABC metric can also be used to compare the task specificity of AOs implemented in commonly used orchestration frameworks\footnote{Implementation details on GitHub in the ComplexScenario folder}. Due to the use of several simple simulators, we did not calculate correctness, reactivity, and traceability for CN and COO4.

\section{Related Work}
\label{sec:relwork}

The majority of ABPM approaches focus on agentic process modeling and mining, e.g., \cite{DBLP:conf/icsoc/NivonS24,klievtsova_conversational_2026,DBLP:conf/ijcai/KouraniB0A24,DBLP:conf/er/ShenPLK24,DBLP:conf/bpm/TourPKS23,DBLP:conf/bpm/MontiLMMR24}, and are complementary to this work. Only few ABPM approaches address the realization and implementation perspective of agentic orchestrations focusing instead on high level architectures such as~\cite{DBLP:conf/webi/KampikMF19}. In \cite{DBLP:conf/kdd/GuanWC0NSZ24}, the authors propose a chain instructor that creates a sequence of steps for a given task to be executed by an agent. This is related to OO2.~\cite{DBLP:journals/emisaij/FettkeKF25} explore the applicability of conceptual modeling techniques for agent orchestrations via several (partially implemented) scenarios which can be classified according to our framework. \cite{DBLP:conf/bpm/KaltenpothSMB25} extract a set of rules from text to augment RPA tasks (OO2 or OO3). By contrast, commercial tools such as Langchain, 
CrewAI, 
Agents SDK, 
Atomic Agents, 
Pydantic AI, and 
Agent Development Kit  have focused on the implementation of agentic orchestrations. In an extensive survey of orchestration frameworks, \cite{fi18060326} present decision criteria for choosing the right orchestration framework, considering ABPM-style frameworks as hierarchical frameworks suitable in scenarios with several agents and a known structure at design time.

\section{Conclusion}
\label{sec:summary}
According to Gartner\footnote{\url{https://tinyurl.com/mrxkr639}} \textsl{``[o]ver 40\% of agentic AI projects will be canceled by the end of 2027, due to escalating costs, unclear business value or inadequate risk controls''}. Hence, the systematic assessment of agentic orchestration options is indispensable for choosing the right implementation for a given scenario. This work provides a first framework with properties and related metrics. Their feasibility has been demonstrated on two real-world scenarios and two implementation frameworks. 
In future work, we will complement the framework with metrics for human involvement and test it for more real-world applications. Moreover, we will investigate other process modeling paradigms such as object-centric, declarative, and hybrid.  

%
%
%

%

\end{document}